# Healthy Harvests: A Comparative Look at Guava Disease Classification Using InceptionV3

Samanta Ghosh
*Dept. of CSE*
East West University
Dhaka, Bangladesh
Samantaewu28@gmail.com

Shaila Afroz Anika
*Dept. of CSE*
East West University
Dhaka, Bangladesh
anikaafroz2002@gmail.com

Umma Habiba Ahmed
*Dept. of CSE*
East West University
Dhaka, Bangladesh
ummahabibaahmed12@gmail.com

B. M. Shahria Alam
*Dept. of CSE*
East West University
Dhaka, Bangladesh
bmshahria@gmail.com

Mohammad Tahmid Noor
*Dept. of CSE*
East West University
Dhaka, Bangladesh
tahmidnoor770@gmail.com

Nishat Tasnim Niloy
*Dept. of CSE*
East West University
Dhaka, Bangladesh
nishat.niloy@ewubd.edu

*Abstract*— Guava fruits often suffer from many diseases. This can harm fruit quality and fruit crop yield. Early identification is important for minimizing damage and ensuring fruit health. This study focuses on 3 different categories for classifying diseases. These are Anthracnose, Fruit flies, and Healthy fruit. The data set used in this study is collected from Mendeley Data. This dataset contains 473 original images of Guava. These images vary in size and format. The original dataset was resized to 256×256 pixels with RGB color mode for better consistency. After this, the Data augmentation process is applied to improve the dataset by generating variations of the original images. The augmented dataset consists of 3784 images using advanced preprocessing techniques. Two deep learning models were implemented to classify the images. The InceptionV3 model is well known for its advanced framework. These apply multiple convolutional filters for obtaining different features effectively. On the other hand, the ResNet50 model helps to train deeper networks by using residual learning. The InceptionV3 model achieved the impressive accuracy of 98.15%, and ResNet50got 94.46% accuracy. Data mixing methods such as CutMix and MixUp were applied to enhance the model's robustness. The confusion matrix was used to evaluate the overall model performance of both InceptionV3 and Resnet50. Additionally, SHAP analysis is used to improve interpretability, which helps to find the significant parts of the image for the model prediction. This study purposes to highlight how advanced models enhance reliable solutions for addressing agricultural disease issues.

*Keywords—Guava, Agriculture, Classification, Deep learning, Image detection, XAI*

## I. Introduction

Guava is a widely cultivated superfruit known in South Asia for its rich vitamin C and fiber content and economic importance. It not only contains vital vitamins and minerals but is also considered a pillar of income for a large number of smallholder farmers. However, guava cultivation faces serious challenges due to a range of devastating plant diseases, predominantly Anthracnose and Fruit Fly infestation. Anthracnose is a disease triggered by fungi like Colletotrichum, rots fruit, and causes it to fall early

Detecting diseases in guava is necessary due to food security, farmers' income, and farming sustainability are affected by it. Guava is an important crop for small-scale and mid-sized farmers in many emerging areas like South Asia. A framework was formulated based on image processing and machine learning techniques to automatically identify multiple diseases in guava fruits and leaves [1]. 400 images of guava fruits were collected with four disease types, and each image was labelled with an indicator. Colour space metrics and Local Binary Pattern (LBP) features were extracted and applied to Principal Component Analysis (PCA) to reduce dimensions. Cubic Support Vector Machine (C-SVM), Fine K-Nearest Neighbours (F-KNN), and Bagged Tree were used for Classification, where Bagged Tree achieved maximum accuracy. It focuses on developing a full process including segmentation, feature extraction, and analyzed classifiers to identify the most effective methods.

Basic image processing and machine learning techniques were applied to detect guava leaf diseases [2]. Guava leaf images were captured by a camera and resized. The diseased region was identified by Region Growing Segmentation, and the images evolved into YCbCr and CIELAB color spaces for the affected areas. They prioritized SVM for its strong performance in binary classification, and it can handle complex data.

According to recent findings, for mobile or edge devices, deep learning and model compression techniques were used to discover guava fruit and foliage infections efficiently [3]. EfficientNet was chosen for its great balance between high accuracy and being lightweight. Quantization reduced the model size without losing accuracy and allowed real-time field applications; for this reason, emphasis is placed on it.

The aim of this study is to design a smart system that is able to classify different guava diseases through deep learning and interpretable artificial intelligence tools. Identifying guava diseases through manual inspection demands significant time. To address these challenges, we aspire to build an automated, efficient, and reliable approach to identify diseased guavas using image analysis. We will use InceptionV3 and ResNet50 to sort guava images by whether they have diseases or not. These models are highly effective for identifying high-level features in disease-affected guava images, and they optimize the classification performance by leveraging deep learning techniques. In addition to achieving high accuracy, our objective extends to ensuring that the visibility and explainability of the model's decision-making process are essential. So, we will use SHAP and LIME, two well-known interpretable AI techniques, to show which parts of the image affected the prediction. This approach will help make the system more reliable and allow fast and accurate detection of guava diseases, which leads to better disease prevention and higher crop yields.

Using both models offers a balanced advantage. InceptionV3 provides multi-scale sensitivity, and ResNet50 affirms depth and accuracy. Both models are pre-trained on large datasets like ImageNet, making effective transfer learning, which speeds up training and enhances accuracy even with small agricultural datasets.

Research Questions:
- To what extent do dataset scale and quality affect model accuracy?
- To facilitate real-time guava leaf disease detection, what system adaptations are necessary for field conditions?

## II. RELATED WORKS

We have gathered some related works to better understand them. Comparing these related works with our proposed work can help us make progress.

The study [4] applied deep learning to analyze fundus images for the early diagnosis of ocular diseases. InceptionV3 & ResNet50 models are employed to categorize images into three classes: normal, macular degeneration, and tessellated fundus. The ResNet50 model achieved the highest accuracy, at 93.81%, while the InceptionV3 model obtained 91.76%. The analysis proposes that this automated system can aid in limiting inaccurate diagnoses caused by factors such as poor image quality & variations in medical expertise.

Fakhrabadi et al. [5] develop a deep learning system to determine COVID-19 severity from lung CT scans. The research used hybrid Inception-ResNet models with dilated convolutions to extract image features and analyze achieved 96.40% accuracy. Outcomes show that dilated residual network performed better than non-dilated networks by collecting more global data. This process proposes precise tool to measure lung involvement for clinical decisions and finds disease progression.

This research explores the application of deep learning models, explicitly ResNet50 & InceptionV3, for defect detection in 3D food printing [6]. This research focuses on the challenges posed by high viscosity materials, which complicate the defect detection process. The ResNet50 model achieved a remarkable accuracy of 93.83%, outperforming InceptionV3, which accuracy 84.62%. The study indicates the importance of model evaluation metrics such like accuracy, precision, recall and F1 scores, which were computed using a confusion matrix to access performance.

The revised ResNet50 model suggested in this study [7] aims to address the limitations by implementing adaptive learning rates and regularization techniques. The results mention that the revised model achieves a training accuracy of 83.95% and a testing accuracy of 74.32%, demonstrating improved reliability in DR detection.

The research highlights ResNet50's ability to mitigate the vanishing gradient problem [8], enhancing feature extraction capabilities. The study also mentions that the integration of multi-model data for upgraded soil analysis and highlighting that ResNet50 achieved an accuracy of almost 95% in various task, showing its probable in precision agriculture and soil monitoring.

This paper investigates [9] the application of deep learning techniques, specifically using the ResNet50 model, for the classification of colorectal cancer. The study demonstrates that ResNet50 outperforms ResNet18, achieving an accuracy of 85% when trained on colon gland image. The research highlights the model's effectiveness in distinguishing between benign and malignant cases, showcasing its potential in medical image analysis.

## III. METHODOLOGY

This Fig. 1 shows the whole process of methodology, designed to develop an explainable image-based Machine learning model. In the beginning of the process, we collect a dataset from Mendeley [10]. The process starts with loading the dataset into the system. These images are preprocessed using CutMix and MixUp. This process enhances the generalization ability of the model. According to preprocessing, the dataset is split into training, validation and testing sets. This dividing process is done for the proper evaluation of model performance. In the next step, multiple machine learning models are trained on the prepared data. After the evaluation is done, the model which gives the most accurate output is selected. In the final stage, some Explainable AI techniques like Grad-CAM are used to visualize the influencing parts of the image which influence the prediction of the image. While visualizing, these techniques ensure transparency and interpretability.

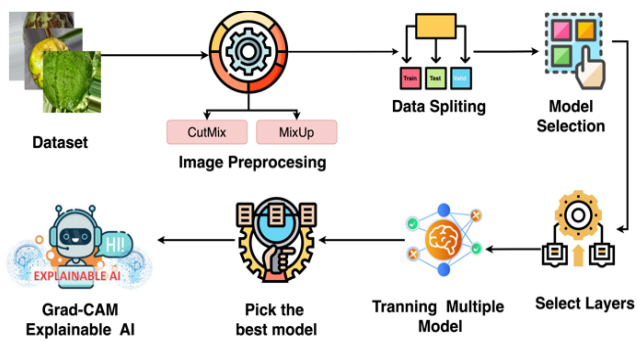

Fig. 1. Methodology

In our study, we used the Guava Fruit Disease Dataset shown in Fig. 2, which is available on Mendeley Data. This dataset consists of 473 images of guava fruits. All images were resized to 256×256 pixels with RGB color mode. Preprocessing techniques were applied to improve the image quality. To increase the number of images (3784), the data preprocessed images are increased. This dataset includes three main classes- Anthracnose, Fruit flies, and Healthy fruits, which are very common conditions of guava farming. Anthracnose is a fungal disease. This disease affects leaves, stems, and fruits. On the other hand, fruit flies happen when insects lay eggs in fruits, and their larvae feed on the fruit flesh. It causes internal damage and makes it unfit for consumption. Lastly, Healthy fruits represent the normal condition of a fruit, which is fit for consumption.

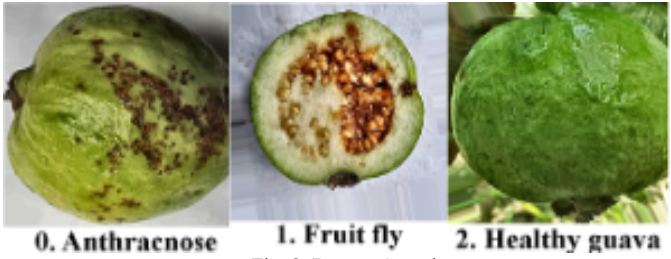
Fig. 2. Dataset Sample

## A. InceptionV3

For obtaining diverse features efficiently, an inception model could be the best option. This model applies multiple convolutional filters of different sizes in parallel within the same layer. It reduces computational cost by using 1x1 convolutions that decrease dimensionality. It also replaces fully connected layers with global average pooling, leading to minimizing the parameters and more efficient performance.

Fig. 3 represents the architecture of the inception model, which enhances CNN performance by capturing feature extraction efficiently. First, the image enters the input layer and passes through convolutional layers that detect shallow features like edges, outlines, and textures. These feature maps are then reduced in size through pooling layers, which improves computational efficiency. The processed feature maps are then passed through inception blocks, which perform parallel convolutions of varying filter sizes, such as 1x1, 3x3, and 5x5 in parallel. The output of these convolutions is combined to form a deep, multi-scale representation of the input. After this, global average pooling is applied to each feature map, where the average value of each map is calculated. In the final stage, the processed features are passed through a SoftMax function. It computes the probability of each class. The class corresponding to the highest score is chosen as the predicted label. The output layer provides the final classification result for the image.

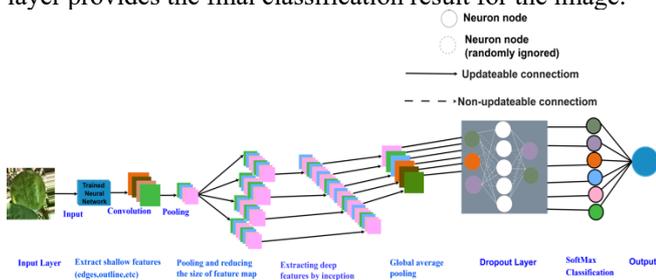
Fig. 3. InceptionV3 Architecture

## B. ResNet50

ResNet50 is a deep convolutional neural network. It consists of 50 layers. It is designed to solve the degradation problem in deep neural networks through residual learning. It is commonly used for image classification and recognition.

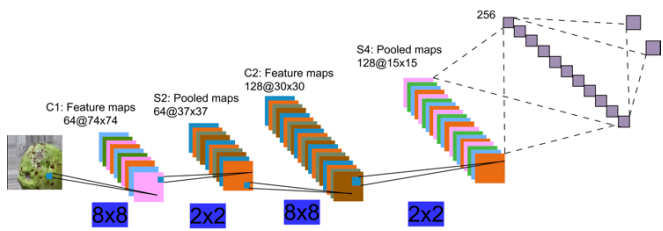
Fig. 4. ResNet50 architecture

This Fig. 4 represents the workflow of the ResNet50 model. This is a deep convolutional neural network model with 50 layers. It is applied for histopathological image analysis. Initially, the input passes through the c1 layer. This layer extracts basic features with 64 filters and produces feature maps of size 74x74. In the pooling layer, this size is reduced to 37x37. After that, another layer that has 128 filters captures more details about the features. This generates a map size of 30x30. Again, in the pooling layer, these are reduced to 15x15. In the final state, the output from the feature extraction layers is passed into a fully connected layer for classification. To make the model train better and faster, this model uses residual connections.

## IV. AUGMENTED REGULARIZATION

### A. CutMix

CutMix is a Data augmentation technique. Fig. 5 shows CutMix adjust a patch of one image with a patch from another. It's a weighted sum based on the patch area .For the input and label denoted as a and b, CutMix is defined as :

$$a\_new = M \odot x\_i + (1-M) \odot x\_j$$
$$b\_new = \partial b\_i + (1-\partial) b\_j$$

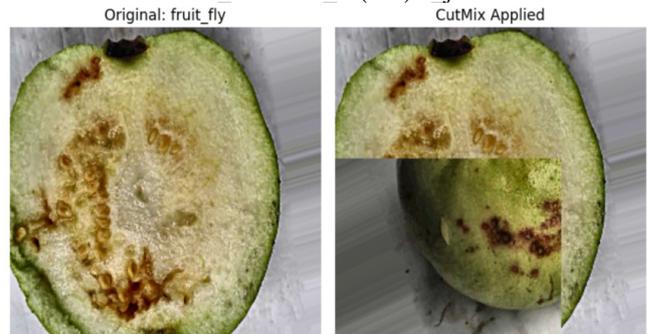
Fig. 5. CutMix image for fruit flies = 0.4

### B. MixUp

Fig. 6 MixUp blends two images linearly and combines both pixel values and labels. In this case, randomly selected images a_i and a_j. Corresponding labels will be y_i and y_j. CutMix defined as:

$$a\_new = \partial a\_i + (1 - \partial) a\_j$$
$$b\_new = \partial b\_i + (1 - \partial) b\_j$$

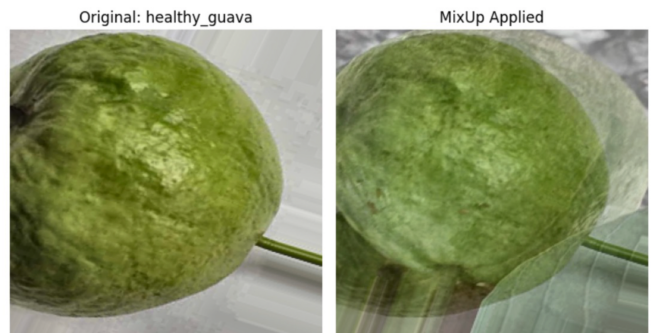
Fig. 6. MixUp image for healthy guava = 0.3

## V. EXPERIMENTAL ANALYSIS

### A. InceptionV3

The Fig. 7 'Training and Validation Accuracy' and Training and Validation Loss' of the InceptionV3 model illustrates excellent training performance. In the first few epochs, the training accuracy reaches nearly 100% with a smooth decreasing training loss. This means that from the training data the model is learning very well. On the other hand, validation accuracy improves quickly from 68% to 85%. The validation loss curve goes up and down after the early epochs. The model learns the training data well, but it cannot generalize effectively. These differences indicate overfitting and making it difficult to work with new data for the model.

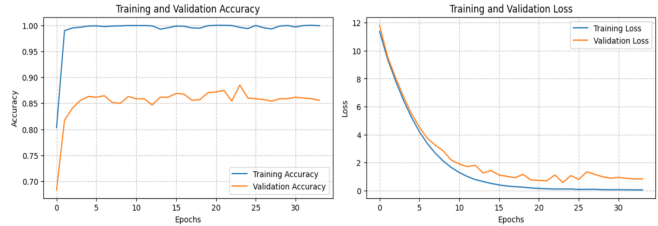

Fig. 7. a) Accuracy for training and validation. b) Loss for training and validation

The loss values over 40 epochs are tracked in "Training and Validation Loss" in Fig. 7. The blue line represents training loss, and the orange line represents validation loss. Initially, the training loss is high (around 14) but reduces quickly within the first 10 epochs and then levels off near 0. This means the model quickly refines its parameters and reduces errors on the training data. The validation loss starts by decreasing too, but remains mildly higher than the training loss, and stays largely unchanged after 10 epochs. Since the two lines stay close and at low values, the model isn't experiencing significant overfitting.

Fig. 8 below represents the confusion matrix of the InceptionV3 model. This confusion matrix shows the model's classification performance among three different classes (Anthracnose, Fruit Fly and Healthy). The number of matrix cells reflects how many samples were classified into each specific pair of actual and predicted labels. The model correctly identifies both Anthracnose (170) and Fruit Fly (118) samples with no misclassification. However, there is a minimal confusion between healthy and fruit fly classes. The model misclassified 7 healthy fruit samples as fruit fly and 84 samples were correctly identified as healthy fruits. Overall, the model is performing well. This model identifies the majority of samples (372 samples were correctly classified out of 379 samples), which indicates that the model is highly effective.

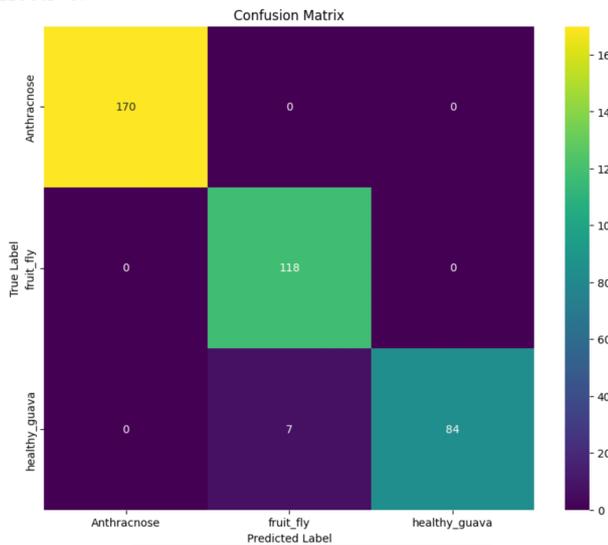

Fig. 8. Confusion Matrix of InceptionV3

The hyperparameters of the InceptionV3 model are shown in a table. In our study, we illustrate each hyperparameter that is used for the InceptionV3 model in Table 1. Our dataset contains three classes, in which the image size is 256x256 pixels. The batch size is 32. There are three color channels of RGB. This model is trained with 100 epochs. This model also used many callbacks such as early stopping, model checkpointing, and a learning rate of 0.0001. The patience is set to 10 and also uses Adam optimizer. By epochs of 6, this model got validation accuracy of 99.96% and validation loss improvement from 5.50 to 4.52. Rescale factor of 1/255 is also applied in the data augmentation parameters and dataset splits, 20% for validation.

TABLE I HYPERPARAMETER TUNING OF RESNET50

| Batch size | 32 | Loss function | categorical_crossentropy |
|---|---|---|---|
| Learning rate | 0.0001 | Number of epochs | 34 |
| Optimizer | Adam | Patience | 10 |

### B. ResNet50

Fig. 9 (a) represents the training and validation accuracy of the ResNet50 model. This illustrates the overall model performance over 60 epochs. The blue line shows the accuracy during the training phase. It increases gradually and remains above 90%. This indicates that the model learns very well from the training data. Validation line accuracy goes up to 80%. But this validation fluctuates continuously and does not achieve a steady phase throughout the process. Sometimes it drops below 40%. That means the model does not work well with the new data. It causes models to overfit.

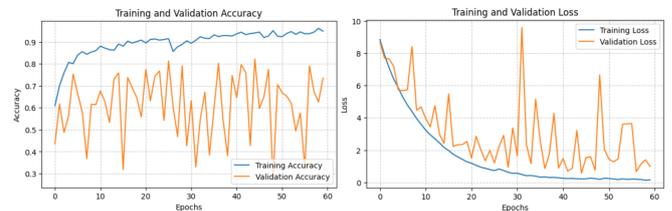

Fig. 9. a) Accuracy for Training and Validation. b) Loss for Training and Validation

Fig. 9 indicates that the training and validation loss curves are over 60 epochs in this figure. The training loss line goes down smoothly. This indicates that the model learns well from the training data. But the validation loss curve shows fluctuating lines. It shows many spikes. This means that with the validation data, the model is detected well.

The confusion matrix represents the model classification performance for ResNet50 among three classes in Fig. 10. These are Anthracnose, Fruit Fly, and Healthy. The model correctly identified 166 samples as Anthracnose. Only 4 samples were misclassified as Healthy Guava. In the Fruit Fly class, 106 samples are correctly identified. But this model misclassified 10 samples as Anthracnose and 2 samples as Healthy Guava. However, the model correctly identified 86 samples as healthy guava out of 91 and misclassified only 3 samples as Fruit Fly and 2 samples as Anthracnose. 377 samples are used in the evaluation. Despite some misclassification, the model shows strong performance, with most of the samples being correctly classified.

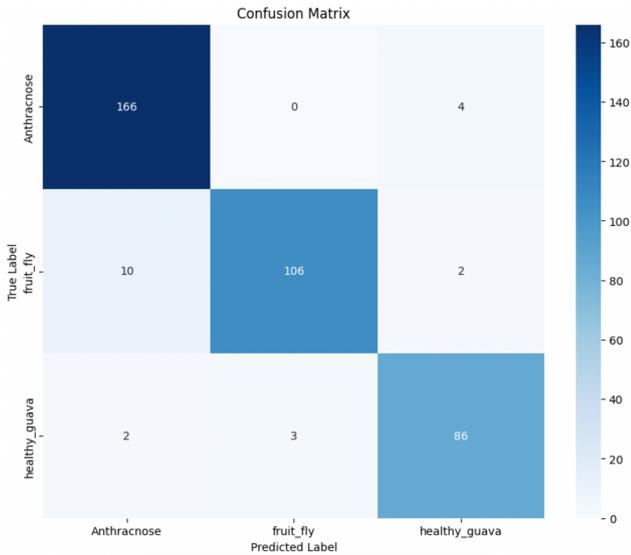

Fig. 10. Confusion Matrix of ResNet50

In Table 2, we have shown the hyperparameters that is used in the ResNet50 model. In this model, the image size is set to 256x256 pixels, 3 channels (RGB), and a batch size of 32. This model also used the Adam optimizer with a learning rate of 0.0001. This model sets 100 epochs for training. However, early stopping is used with patience 15. This model stops training after 7 epochs without any improvement in validation loss. The validation results got an accuracy of 83.41 and a validation loss of 0.6790

TABLE II HYPERPARAMETER TUNING INCEPTIONV3

| Batch size | 32 | Loss function | categorical_crossentropy |
|---|---|---|---|
| Learning rate | 0.0001 | Number of epochs | 60 |
| Optimizer | Adam | Patience | 15 |

The effect of MixUp and CutMix augmentations on InceptionV3 model performance is shown in Table 3. These techniques are applied to improve model performance. Alpha MixUp and Alpha Cutmix help to get the best performance for both Validation Accuracy and Validation Loss. We get the highest accuracy of 85.85% and the lowest validation loss value of 1.54 when the Alpha MixUp value is 0.25 and the Alpha CutMix value is 0.4. However, increasing more Alpha values may cause low accuracy and increase validation loss. Moderate augmentations are good for improved performance.

TABLE III HYPERPARAMETER TUNING

| Alpha MixUP | Alpha CutMix | Validation Accuracy (%) | Validation Loss |
|---|---|---|---|
| 0 | 0 | 85.37 | 2.4724 |
| 0.2 | 0.3 | 85.83 | 1.8474 |
| 0.25 | 0.4 | 85.85 | 1.5442 |
| 0.3 | 0.5 | 85.19 | 2.3139 |
| 0.35 | 0.6 | 84.96 | 2.3728 |

VI. EXPLAINABLE AI

A. SHAP

This Fig. 11 presents the damaged part of the cross-section of the guava fruit. SHAP mainly highlights important features and assigns values to the feature. This also explains how much the feature impacts the model's decision. Here, GRAD-CAM is a prominent feature. The right image is the output of SHAP GRAD-CAM. It illustrates the heatmap overlay created by the InceptionV3 model. This highlights the most influential part that contributed significantly to the model's prediction. The picture shows that the heatmap aligns with the damaged parts. That means the model correctly identifying the areas as important features.

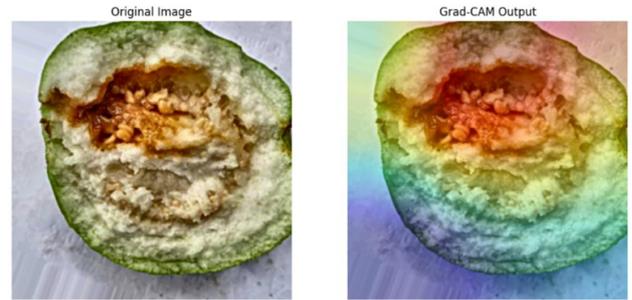

Fig. 11. Guava Disease Classification using SHAP

B. LIME

Fig. 12 shows a LIME Visualization of the InceptionV3 model. This visualization is used to highlight the model predictions of the most important regions in the image. The original image is cut open, Guava. The Yellow line of the lime explanation area indicates the most important region that influences the model's decision. The last one is the heatmap explanation, where the highlighted areas are the most important for the prediction. Blue areas are less important.

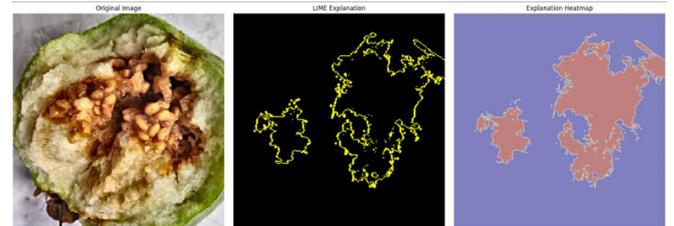

Fig. 12. Guava Disease Classification using LIME

VII. COMPARISON

Table 4 presents a comparison of performance between the InceptionV3 and ResNet50 models. The InceptionV3 model achieved a higher accuracy of 98.15%, where ResNet50 achieved 94.46%. According to all evaluating metrics (Precision, Recall, and F1-score) and accuracy, the overall performance of the InceptionV3 model is better than ResNet50 architecture. This indicates that InceptionV3 provides more robust and consistent performance for the given dataset.

TABLE IV COMPARISON BETWEEN THE TWO ARCHITECTURES

| Model Name | Classes | Precision | Recall | F1-score | Accuracy |
|---|---|---|---|---|---|
| InceptionV3 | Anthracnose | 1.00 | 1.00 | 1.00 | 0.9815 |
| | Fruit flies | 0.94 | 1.00 | 0.97 | |
| | Healthy | 1.00 | 0.92 | 0.96 | |
| ResNet 50 | Anthracnose | 0.93 | 0.97 | 0.95 | 0.9446 |
| | Fruit flies | 0.97 | 0.89 | 0.93 | |
| | Healthy | 0.93 | 0.94 | 0.93 | |

Fig. 13 compares the Training accuracy and validation accuracy of the two models. Here, the training accuracy of the InceptionV3 model is 98%, where the ResNet50 model achieved 94%. That means the InceptionV3 model trained better than the ResNet50 model. The validation accuracy of the InceptionV3 model (88%) is higher than ResNet50 model (68%). That means this Inception Model performs better than the ResNet50 model.

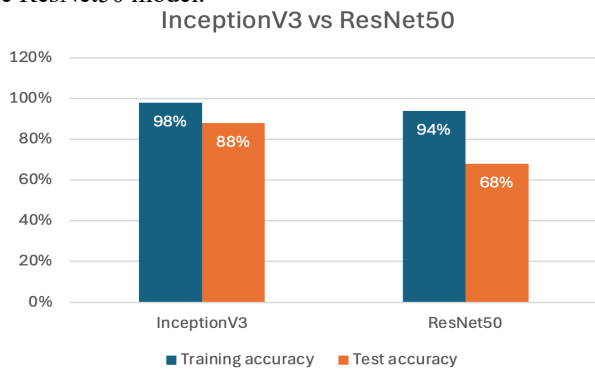

Fig. 13. Accuracy Comparison Between the Two Architectures

Table 5 shows the comparison of the performance of deep learning model which are applied in previous studies alongside the proposed model. It shows the comparison based on the applied methods and their accuracy. Mawardi et al. [6] achieved 93.83% accuracy using ResNet50 and 84.62% accuracy using InceptionV3. Using the same models, Yuhang Pan et al. [4] achieved lower accuracies. He achieved 3.81% for ResNet50 and 91.76% for InceptionV3. Chun-Ling Lin and Kun-Chi Wu [7] achieved a training accuracy of 83.95% and a test accuracy of 74.32% using a revised version of ResNet50. On the other side, the proposed model better accuracy. It achieved 98.15% accuracy using InceptionV3 and 94.46% using ResNet50. These results are proving that the proposed model is more effective than the existing methods.

TABLE V COMPARISON WITH PREVIOUS WORKS

| Author | Method | Accuracy (%) |
|---|---|---|
| Mawardi et al. [6] | ResNet50 | 93.83 |
| | InceptionV3 | 84.62 |
| Yuhang Pan et al. [4] | ResNet50 | 93.81 |
| | InceptionV3 | 91.76 |
| Chun-Ling Lin [7] | ResNet50 | 74.32 |
| Proposed Model | ResNet50 | 94.46 |
| | InceptionV3 | 98.15 |

## VIII. CONCLUSION

Guava is a common fruit all over the world. A huge amount of guava is produced every day. Besides, guava is affected by many diseases. This study was done to train a model to detect efficiently the disease of guava fruit. To classify the diseases, three different categories are focused on in the study. The categories are Anthracnose, Fruit flies, and Healthy fruit. A dataset containing 473 images of guava was used. In this study, two different deep learning models, InceptionV3 and ResNet50, are implemented. This MODEL significantly improved model performance. The Inceptionv3 model achieved 98.15% accuracy, and the ResNet50 model achieved 94.46% accuracy. The whole output of this study is comparatively better than the previous studies. To provide accurate solutions for detecting guava diseases, this study highlights the potential of applying machine learning. There are some limitations of the study. The data used in the study could have more sample images. If the number of sample images were more than we used in this study, the output would be more generalized and accurate. Also, the sample images that were used were not clear enough. If the images were more standard, the result could have been more accurate. So, there is a further scope of study to get more generalized and accurate results. This model could be a solution to the problem if it is applied in the right way.